\pdfoutput=1

\documentclass[11pt]{article}

\usepackage[review]{ACL2023}

\usepackage{times}
\usepackage{latexsym}
\usepackage{graphicx}
\usepackage{bm}
\usepackage{multirow}
\usepackage{subfigure}[caption=false]
\usepackage{tablefootnote}

\usepackage{algorithm}
\usepackage{algorithmicx}
\usepackage{algpseudocode}
\usepackage{amsmath}

\algrenewcommand\algorithmicindent{1em}%

\usepackage[T1]{fontenc}

\usepackage[utf8]{inputenc}

\usepackage{microtype}

\usepackage{inconsolata}

\title{Fast Rule-Based Decoding: Revisiting Syntactic Rules in Neural Constituency Parsing}

\author{Tianyu Shi, Zhicheng Wang, Liyin Xiao, Cong Liu\thanks{\ \ Corresponding Author}\\
  School of Computer Science and Engineering, Sun Yat-sen University, China\\
  {\tt \{shity3, wangzhch23, xiaoly28\}@mail2.sysu.edu.cn} \\
  {\tt liucong3@mail.sysu.edu.cn} }

\begin{document}
\maketitle
\begin{abstract}
Most recent studies on neural constituency parsing focus on encoder structures, while few developments are devoted to decoders. Previous research has demonstrated that probabilistic statistical methods based on syntactic rules are particularly effective in constituency parsing, whereas syntactic rules are not used during the training of neural models in prior work probably due to their enormous computation requirements. In this paper, we first implement a fast CKY decoding procedure harnessing GPU acceleration, based on which we further derive a syntactic rule-based (rule-constrained) CKY decoding. 
In the experiments, our method obtains 95.89 and 92.52 F1 on the datasets of PTB and CTB respectively, which shows significant improvements compared with previous approaches. 
Besides, our parser achieves strong and competitive cross-domain performance in zero-shot settings.
\end{abstract}

\begin{figure}[t]
\centering
\subfigure{
    \begin{minipage}{\linewidth}
    \centering
    \includegraphics[width=\linewidth]{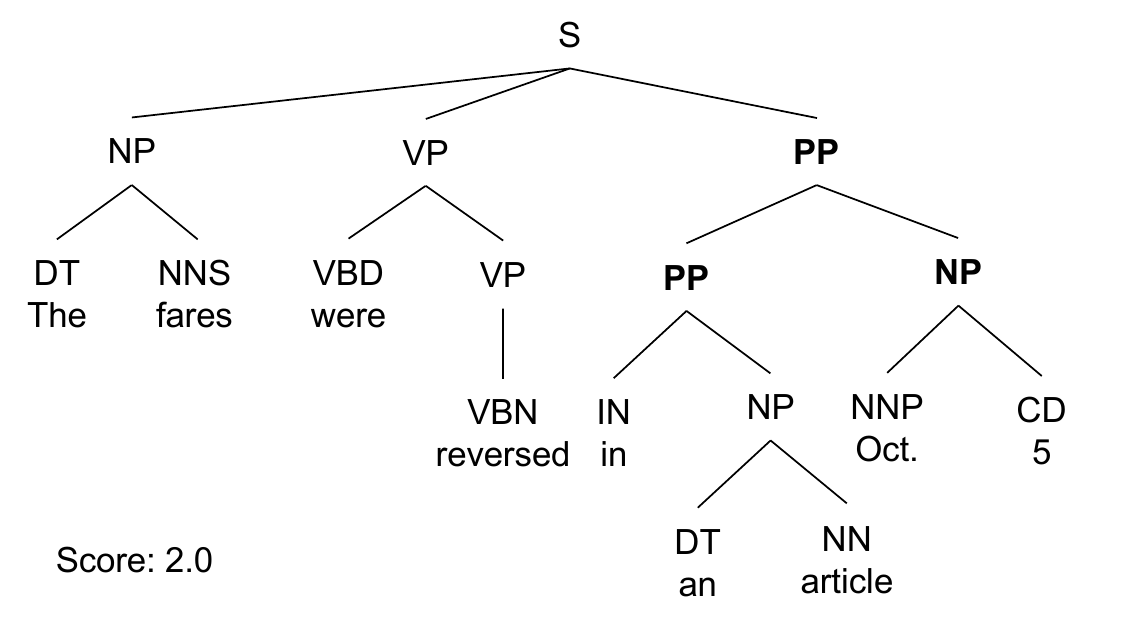}
    \caption*{(a) Conventional CKY Decoding}
    \end{minipage}
}
\subfigure{
    \begin{minipage}{\linewidth}
    \centering
    \includegraphics[width=\linewidth]{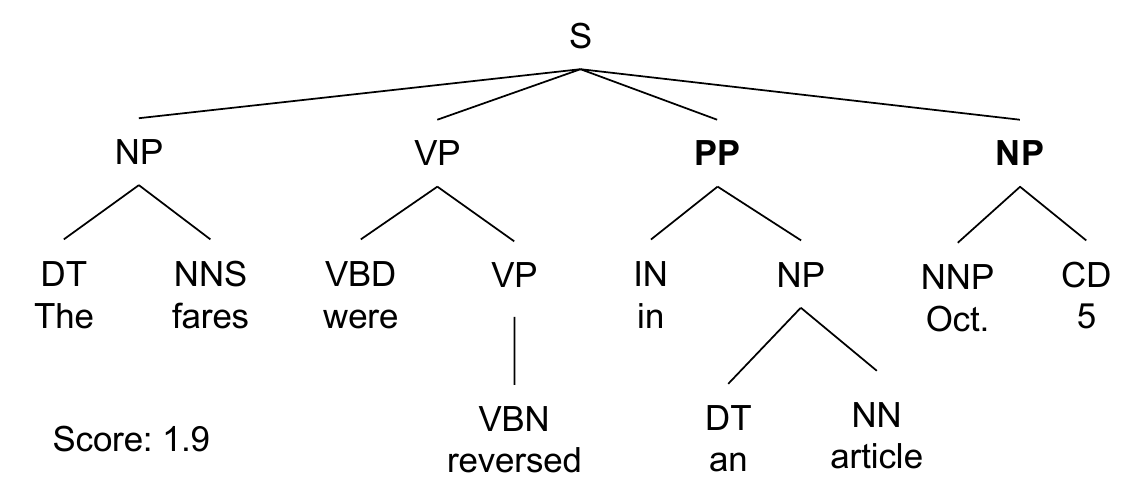}
    \caption*{(b) Rule-Based CKY Decoding}
    \label{1b}
    \end{minipage}
}
\caption{An example of conventional CKY decoding and rule-based CKY decoding. Although the conventional CKY decoding obtains a higher score, PP $\rightarrow$ PP NP does not conform to any syntactic rules, resulting in a wrong parse tree.}
\label{fig:te}
\end{figure}

\section{Introduction}

Constituency parsing is a fundamental task in natural language processing that involves assigning constituent labels to each span of a sentence and constructing a parse tree. 
Constituency parsing is useful for a variety of downstream NLP tasks, including information extraction, machine translation, and question answering.

There are mainly two paradigms for constituency parsing, transition parsers and chart parsers. In the former, a constituent parse tree is constructed by a sequence of shift-reduce operations \citep{dyer-etal-2016-recurrent}. Although transition parsers have $O(n)$ time complexity, they usually require additional feature designs \cite{liu-zhang-2017-shift} to increase the accuracy. 
In the latter, the decoder utilizes the CKY algorithm \cite{1966,1967,1970} to obtain the optimal parse tree based on the span scores assigned by neural networks.
In this paper, we concentrate on chart parsers, which have the advantage of achieving globally optimal solutions, rather than being limited to local optima as in transition parsers.

In the past, the CKY algorithm constructs a parse tree using context-free grammar (CFG), which contains a large number of syntactic rules. 
Probabilistic context-free grammar (PCFG) assigns a probability to each rule based on CFG, and the probability of a parse tree is the product of probabilities of all the rules applied when constructing the tree, where the parse tree with the maximum probability indicates the optimal tree.

In recent studies, most researchers utilize neural models for constituency parsing. The seq2seq model \cite{grammar2015} interprets parsing as a sequential task, and when the attention mechanism \cite{attention2014} is employed, it requires $O(n^2)$ complexity. \citet{stern-etal-2017-minimal} present a basic neural structure that independently scores the label for each span. Rather than depending on grammar to construct the parse tree, this approach enables the neural network to learn implicit rules. Subsequent work \cite{kitaev-klein-2018-constituency,YuZhang2020FastAA,wei-etal-2020-span} has achieved outstanding performance by improving model structures or span representations.

The limitation of previous decoding methods is that the highest-scoring label is first determined for each span, and then the CKY algorithm is used to locate the optimal split point of the span, starting from the bottom.
However, if the labels of different spans have already been decided, the correlation between labels and spans will be ignored throughout the parse tree constructing process, as the combined spans with different labels must be consistent with syntactic rules.
As shown in Figure \ref{fig:te}, although the conventional neural CKY decoding obtains a higher score, it constructs a wrong tree.

To attack this problem, we propose a rule-based CKY decoding algorithm. 
The main idea is to ensure that the label of each node in the tree and those of its children must be combined into one of the rules that exist in the training data. 
The main difference between our decoding and conventional neural CKY decoding is that our decoding determines a per-label optimal split point for each span, which is constrained by the rules, while conventional decoding determines the label of each span and its split point independently.

The purpose of our rule-based decoding goes beyond improving decoding performance. We implement a fast GPU accelerated version of our decoding so that we can use it in training without incurring a significantly prolonged training time. In contrast to the conventional CKY decoding written in \emph{C} of $O(n^3)$ complexity on sentence length $n$, our fast rule-based algorithm is $O(n)$ in terms of GPU operations, although its theoretical complexity is $O(n^3|\mathtt{L}||\mathtt{R}|)$, where $|\mathtt{L}|$ is the number of labels and $|\mathtt{R}|$ is the number of rules. Compared with another of our implemented GPU accelerated CKY decoding, the actual training time of our GPU accelerated rule-based decoding merely increases each training epoch from 15 minutes to 17 minutes on a GeForce Titan RTX.

To evaluate the proposed method, we build our parser based on a prior state-of-the-art parser \cite{mrini-etal-2020-rethinking}, and evaluate on PTB \cite{marcus-etal-1993-building} and CTB \cite{ctb}. 
Furthermore, we test the generalization ability of our parser on MCTB \cite{yang-etal-2022-challenges} in zero-shot settings. 
The experimental results show that our rule-based CKY decoding algorithm brings significant improvements to constituency parsing in both English and Chinese. In conclusion, the main contributions of our research can be summarized as follows:

\begin{itemize}
\item We propose a GPU accelerated rule-based constituency decoder, whose complexity is $O(n)$ in terms of GPU operations.
\item With our accelerated rule-based decoder, we investigate the joint training that jointly optimizes the losses from a rule-based parser and a conventional rule-less parser.
\item Experimental results show that our method outperforms previous approaches, which achieves 95.89 and 92.52 F1 on PTB and CTB respectively, and also achieves strong cross-domain performance in zero-shot settings.
\end{itemize}

\section{Related Work}
\label{relate}

\noindent \textbf{Label parsing} 
\ Recently, numerous methods have been presented to simplify constituency parsing to a sequence tagging task.
Supertagging \cite{bangalore-joshi-1999-supertagging} is one of these methods, which is common in CCG parsers.
CCG parsers impose constraints on the valid derivation of supertags and require a complex search procedure to find a sequence with the highest score.
\citet{gomez-rodriguez-vilares-2018-constituent} propose a method with an infinite size label set, which completely models parsing as tagging. 

\

\noindent \textbf{Transition parsing} 
\ Transition parsers utilize a transformation system \cite{henderson-2003-inducing, sagae-lavie-2005-classifier} that takes a sequence of lexical inputs and produces a sequence of tree-constructing actions.
They parse sentences sequentially and have linear time complexity.
Using special features \cite{gnn}, or advanced decoding methods \cite{cross-huang-2016-span}, the accuracy of transition-based parsers is improved effectively.
\citet{kitaev-klein-2020-tetra} use only four actions, further improving the efficiency.

\begin{figure*}[t]
\centering
\begin{tabular}{c}
\includegraphics[width=\linewidth]{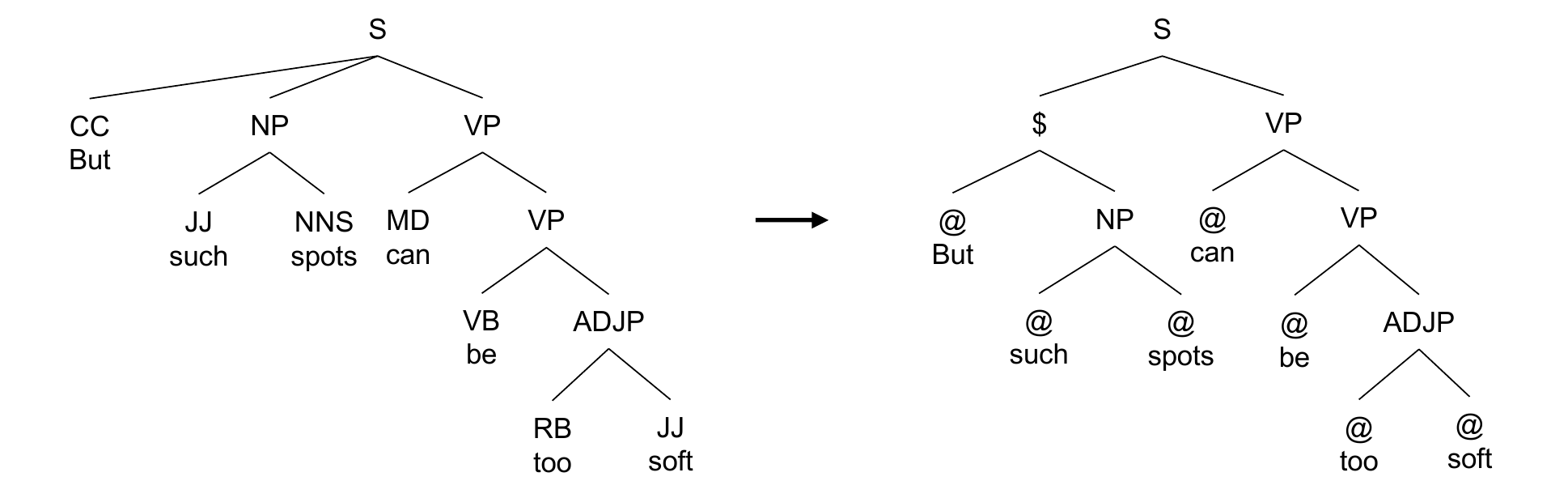}
\end{tabular}
\caption{An example of converting an $n$-ary parse tree to a binary parse tree. }
\label{fig:trees}
\end{figure*}

\begin{algorithm}[t]
\caption{Conventional Neural CKY Algorithm} 
\begin{algorithmic}
\Require Sentence length $n$. Span score $s[i,j,\ell]$, for $0\le i<j\le n$ and $\ell \in \mathtt{L}$, $\mathtt{L}$ being the label set.
\Ensure Sub-tree score $t[i,j]$ for every span $(i,j)$. Left-right children split point $K[i,j]$ for each of the above sub-trees.
\For{$i = 0$ \rm to $n-1$}
    \State $t[i,i+1] = \mathop{\rm max}\limits_{\ell \in \mathtt{L}}\{s[i,i+1,\ell]\}$
\EndFor
\For{$ss = 2$ \rm to $n$} \Comment{span size}
    \For{all spans $(i,i+ss)$}
        \State $k= \mathop{\rm argmax}\limits_{i<j<i+ss}\{t[i,j]+t[j,i+ss]\}$
        \State $K[i,i+ss] = k$
        \State $t[i,i+ss] = \mathop{\rm max}\limits_{\ell \in \mathtt{L}}\{s[i,i+ss,\ell]\}+t[i,k]$
        \State ~~~~~~~~~~~~~~~~~~~~~~$+t[k,i+ss]$
    \EndFor
\EndFor
\end{algorithmic}
\label{a1}
\end{algorithm}

\noindent \textbf{Chart parsing} 
\ Chart parsers use the CKY algorithm \cite{1966,1967,1970} to predict the optimal parse tree that has the highest score.
Recent chart parsers are based on neural networks \cite{kitaev-klein-2018-constituency,zhou-zhao-2019-head}, whereas they sometimes produce the optimal tree that does not conform to the syntactic rules, resulting in a wrong tree.
In contrast, our method uses syntactic rules extracted from the training set to improve the parsing accuracy.

\begin{algorithm}[t]
\caption{Fast Neural CKY Algorithm} 
\begin{algorithmic}
\Require Sentence length $n$. Span score $s[i,j,\ell]$, for $0\le i<j\le n$ and $\ell \in \mathtt{L}$, $\mathtt{L}$ being the label set.
\Ensure Sub-tree score $t[i,j]$ for every span $(i,j)$. Left-right children split point $K[i,j]$ for each of the above sub-trees.
\State \Comment{We view 2D as 1D when with indexes $I,L,R$}
\State $I=range(0,n-1)\times n +range(1,n)$
\State $t[I]=s[I,:].max(dim=-1)[0]$
\For{$ss = 2$ \rm to $n$} \Comment{span size}
    \State $I = range(0,n-ss+1)\times n+$
    \State ~~~~~~~$range(0,n-ss+1)+ss-1$
    \State $I = I.add\_dim(-1)$
    \State $L = I+range(1-ss,0).add\_dim(0)$
    \State $R = I+range(1,ss).add\_dim(0)\times n$
    \State $t',K[I] = (t[L]+t[R]).max(-1)$
    \State $t[I] = t[I]+t'$
\EndFor
\end{algorithmic}
\label{a2}
\end{algorithm}

\section{Decoding Algorithms}
In this section, we first introduce how to acquire syntactic rules from the training set.
Then, we present the conventional neural CKY decoding algorithm and a fast GPU accelerated version.
Finally, we describe our sequential and fast rule-based CKY decoding algorithms.

\subsection{Rule Acquisition}
\label{BBB}
Syntactic rules in our method are all derived from the training set of PTB and CTB. 
As shown in Figure \ref{fig:trees}, we first use the left-branching binarization algorithm following the same implementation of \citet{kitaev-klein-2018-constituency}.
We replace all part-of-speech (POS) tags with the same label @ to reduce the number of rules, since they do not contribute much to the parsing performance when contextual embeddings are used \cite{kitaev-klein-2018-constituency}. 
For nodes generated during binarization, we denote them with the same label \$ to prevent the explosion of the number of syntactic rules.
The resultant parse tree in Figure \ref{fig:trees} contains six rules: S $\rightarrow$ \$ VP, \$ $\rightarrow$ @ NP, NP $\rightarrow$ @ @, VP $\rightarrow$ @ VP, VP $\rightarrow$ @ ADJP and ADJP $\rightarrow$ @ @. 

As an alternative labeling scheme, we denote generated nodes using a concatenation of \$ with the label of its left child. 
For example, \$ $\rightarrow$ @ NP is replaced with \$@ $\rightarrow$ @ NP in our second labeling scheme. These two labeling schemes are compared in Section \ref{sec:com}.

\begin{algorithm}[t]
\caption{Tree Decoding} 
\begin{algorithmic}
\Require Sentence length $n$. Span score $s[i,j,\ell]$, for $0\le i<j\le n$ and $\ell \in \mathtt{L}$, $\mathtt{L}$ being the label set. Left-right children split point $K(i,j)$.
\Ensure The optimal tree $T$.
\State $l=\mathop{\rm argmax}\limits_{\ell\in \mathtt{L}}\{s(0,n,\ell)\}$
\State $T=Tree(span=(0,n),label=l)$
\State $stack=$[]
\State $stack.push(T)$
\While{$len(stack)>0$}
    \State $v=stack.pop()$
    \State $(i,j)=v.span$
    \State $k=K[i,j]$
    \State $l_1 = \mathop{\rm argmax}\limits_{\ell\in \mathtt{L}}\{s[i,k,\ell]\}$
    \State $c_1=Tree(span=(i,k),label=l_1)$
    \State $l_2 = \mathop{\rm argmax}\limits_{\ell\in \mathtt{L}}\{s[k,j,\ell]\}$
    \State $c_2=Tree(span=(k,j),label=l_2)$
    \If{$i+1<k$}
        \State $stack.push(c1)$
    \EndIf
    \If{$k+1<j$}
        \State $stack.push(c2)$
    \EndIf
    \State $v.children=(c1,c2)$
\EndWhile
\end{algorithmic}
\label{a3}
\end{algorithm}

\subsection{Conventional Neural CKY Algorithm}
\label{sec:chart}
We compare the sequential and fast CKY algorithms side-by-side in Algorithm \ref{a1} and \ref{a2}.
As shown in Algorithm \ref{a1}, this method first determines the highest-scoring label for each span, and then uses the CKY algorithm to locate the optimal split point for each span.
Due to its $O(n^3)$ time complexity, it is programmed in \emph{C} and interfaced to \emph{Python} for acceleration.

We implement a fast neural CKY algorithm using parallel operations provided by GPUs.
In Algorithm \ref{a2}, function $range(a,b)$ returns a 1D tensor ranging between integers $a$ and $b$, which is equivalent to $torch.arange$ in \emph{PyTorch}. 
For convenience of presentation, we adopt the \emph{PyTorch} function $a.max(dim=i)$ that returns the maximum values on dimension $i$ of tensor $a$ and the indexes to the maximum values on dimension $i$. 
With this function, we are able to reduce $O(n)$ complexity.
Function $a.add\_dim(i)$ adds an $i^{th}$ dimension in tensor $a$, which is equivalent to $a.unsqueeze(i)$ in \emph{PyTorch}. 
In Algorithm \ref{a2}, $I$ is a 1D vector representing the 1D view of the 2D indexes of all the spans of size $ss$, $L$ is a vector representing indexes of all the left sub-spans of $I$, and $R$ is all the right sub-span indexes of $I$.
Using these indexes, we can reduce another $O(n)$ complexity by computing the spans of the same size in parallel.
The complexity is thus $O(n)$ in terms of GPU operations.

As shown in Algorithm \ref{a3}, the optimal tree can be constructed using $K(i,i+ss)$, which are the optimal left-right children split points for spans $(i,i+ss)$ returned from Algorithm \ref{a1} or \ref{a2}.

\subsection{Rule-Based CKY Algorithm}
\label{sec:rule}
Again, we compare the sequential and fast rule-based CKY algorithms side-by-side in Algorithm \ref{a4} and \ref{a5}.

The sequential rule-based CKY algorithm, as shown in Algorithm \ref{a4}, computes the sub-tree score $t(i,j,\ell)$ with an additional per-label dimension, with which it can be constrained that $t(i,j,\ell)=s(i,j,\ell)+t(i,k,\ell_1)+t(k,j,\ell_2)$ only when $\ell \rightarrow \ell_1 \ell_2$ is a syntactic rule appears in the training set in addition to $i<k<j$.
Since we consider each span in the case of all possible labels and rules, the total time complexity is $O(n^3|\mathtt{L}||\mathtt{R}|)$, where $|\mathtt{L}|$ is the number of labels and $|\mathtt{R}|$ is the number of rules.

The fast rule-based CKY algorithm is listed in Algorithm \ref{a5}, where we pre-compute a tensor $rules$ of size $|\mathtt{L}|\times |\mathtt{R}| \times 3$, which for each label $\ell \in \mathtt{L}$ lists the rules $\ell \rightarrow \ell_1 \ell_2$ that starts with $\ell$. 
The last dimension of tensor $rules$ stores the labels $\ell$, $\ell_1$ and $\ell_2$ of each rule.
Indexes $I$ for the spans of size $ss$, their left children indexes $L$, and their right children indexes $R$ are defined exactly the same as in Algorithm \ref{a2}. 
The result of Algorithm \ref{a5} includes $rule\_per\_split$ applied in each optimal split point in addition to the optimal split points $K(i,i+ss)$.
As a result, the complexity of this algorithm is $O(n)$ in terms of GPU operations.

We omit the decoding algorithm for the rule-based CKY algorithm, which simply extends the CKY decoding in Algorithm \ref{a3}, by using $rule\_per\_split$ to determine the label of each node and its children according to the rule applied at their optimal split points.

\begin{algorithm}[t]
\caption{Sequential Rule-Based CKY Algorithm} 
\begin{algorithmic}
\Require Sentence length $n$. Span score $s[i,j,\ell]$, for $0\le i<j\le n$ and $\ell \in \mathtt{L}$, $\mathtt{L}$ being the label set. Rule set $\mathtt{R}$. 
\Ensure Sub-tree score $t[i,j,\ell]$ for every span $(i,j)$ with label $\ell$. Left-right children split point $K[i,j, \ell]$ for each of the above sub-trees.
\For{$i = 0$ \rm to $n-1$}
    \For{$\ell$ \rm in $\mathtt{L}$}
        \State $t[i,i+1,\ell]=s[i,i+1,\ell]$
    \EndFor
\EndFor
\For{$ss = 2$ \rm to $n$} \Comment{span size}
    \For{all spans $(i,i+ss)$ and $\forall \ell \in \mathtt{L}$}
        \State $k,r= \mathop{\rm argmax}\limits_{\substack {i<j<i+ss\\r=\ell \rightarrow \ell_1 \ell_2\\ r \in \mathtt{R}}}
        \{t[i,j,\ell_1]+t[j,i+ss,\ell_2]\}$
        \State $K[i,i+ss,\ell]=(k,r)$
        \State $t[i,i+ss,\ell] = s[i,i+ss,\ell]+t[i,k,\ell_1]+$
        \State ~~~~~~~~~~~~~~~~~~~~~~~~~~$t[k,i+ss,\ell_2]$
    \EndFor
\EndFor
\end{algorithmic}
\label{a4}
\end{algorithm}

\begin{algorithm}[t]
\caption{Fast Rule-Based CKY Algorithm} 
\begin{algorithmic}
\Require Sentence length $n$. Span score $s[i,j,\ell]$, for $0\le i<j\le n$ and $\ell \in \mathtt{L}$, $\mathtt{L}$ being the label set. $rules\ (|\mathtt{L}|\times |\mathtt{R}|\times 3$) list the rules of each label, where each rule $r$ is represented by three labels.
\Ensure Sub-tree score $t[i,j,\ell]$ for every span $(i,j)$ with label $\ell$. Left-right children split point $K[i,j, \ell]$ for each of the above sub-trees. $rule\_per\_split[i,j,\ell,r]$ returns the best rule $r$ for each of the above split points.
\State \Comment{We view 2D as 1D when with indexes $I,L,R$}
\State $I=range(0,n-1)\times n +range(1,n)$
\State $t[I,:]=s[I,:]$
\For{$ss = 2$ \rm to $n$} \Comment{span size}
    \State $I, L, R = \cdots$\Comment{same as in Algorithm \ref{a2}}
    \State $t_1=t[L,rules[:,:,1]]$
    \State $t_2=t[R,rules[:,:,2]]$
    \State $t'=t[I,:].add\_dim(-1)+t_1+t_2$
    \State $t'', rule\_per\_split[I,:,:]=t'.max(-1)$
    \State $t[I,:], K[I,:]=t''.max(-1)$
\EndFor
\end{algorithmic}
\label{a5}
\end{algorithm}

\section{Syntactic Parsing Model}
\label{method}
\begin{figure*}[t]
\centering
\includegraphics[width=\linewidth]{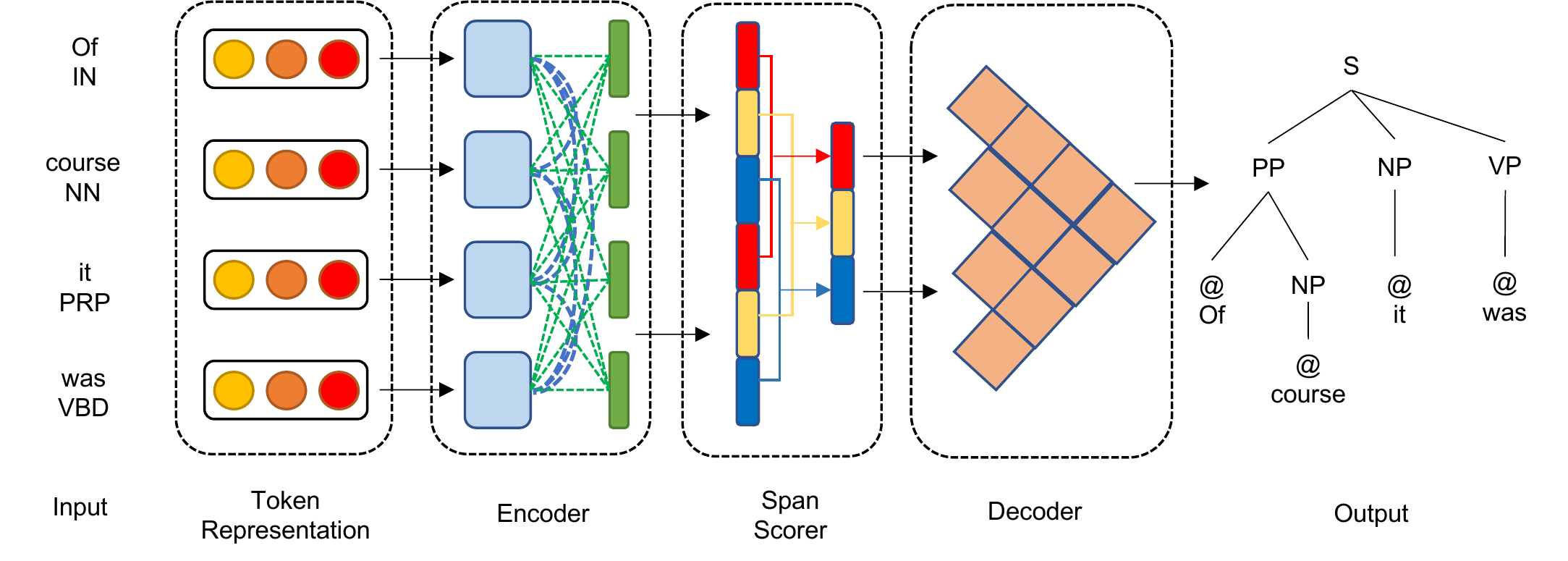}
\caption{The architecture of our syntactic parsing model.}
\label{fig:model}
\end{figure*}

Our parser is built on an encoder-decoder backbone. 
The overall architecture is shown in Figure \ref{fig:model}.

\subsection{Encoder}
The encoder of our model is based on self-attention layers \cite{transfomer}. We follow the attention partitioning of \citet{kitaev-klein-2018-constituency}, which separates content features from position features. After self-attention layers, there is a label attention layer \cite{mrini-etal-2020-rethinking} that uses attention heads to represent labels with potential semantics.

\subsection{Span Scorer}
The span representation follows the approaches of \citet{kitaev-klein-2018-constituency} and \citet{gaddy-etal-2018-whats}. For a span that starts with the $i^{th}$ word and ends with the $j^{th}$ word, the corresponding span vector can be computed as:
$$
s_{ij}=[\overrightarrow{y_j}-\overrightarrow{y_{i}};\overleftarrow{y_{j+1}}-\overleftarrow{y_{i+1}}],
$$
where $\overleftarrow{y_i}$ and $\overrightarrow{y_i}$ are respectively the forward and backward representations of the $i^{th}$ word, by splitting the output vector from the encoder in half.
The span score takes span $s_{ij}$ as the input, which is obtained using a one-layer feed-forward network:
$$
S(i,j)=W_2ReLU(LN(W_1s_{ij}+b_1))+b_2,
$$
where $W_1, W_2, b_1$ and $b_2$ represent learnable parameters, $LN$ represents Layer Normalization. 
The span score with label $\ell$ is the corresponding value of the output vector:
$$
S(i,j,\ell)=[S(i,j)]_\ell.
$$

\subsection{Decoder}
There are two decoders in our model, the conventional CKY decoder and the rule-based CKY decoder.
The score $S(T)$ of the whole parse tree $T$ is to sum the score in each span:
$$
S(T) = \sum_{i,j,\ell\in T}S(i,j,\ell).
$$

The goal of our parser is to obtain the optimal tree 
$T^*$, such that all trees $T$ satisfy the following marginal constraint:
$$
S(T^*) \geq S(T)+\Delta(T,T^*),
$$
where $\Delta$ denotes the Hamming loss for each span. 
The corresponding objective loss function is the hinge loss:
$$
L=\mathop{\rm max}(0,\mathop{\rm max}\limits_{T}[S(T)+\Delta(T,T^*)]-S(T^*)),
$$
where $L$ can represent both the loss $L_c$ from a conventional rule-less parser and the loss $L_r$ from a rule-based parser.

During training, we investigate the joint training that optimizes losses from two kinds of parsers. 
To control the influence of combination, we apply a weight $\lambda$ to obtain the overall loss:
$$
L^* = (1-\lambda)L_c+\lambda L_r,
$$
where $\lambda$ ranges from 0 to 1. 
When $\lambda$ is set between 0 and 1, it indicates that we use two decoders with a mixture of rule-based training and rule-less training.
During inference, we use our rule-based CKY decoder to construct the parse tree.

\section{Experiments}
\label{experiments}
We evaluate our syntactic parsing model on English Penn Treebank (PTB) 3.0 \cite{marcus-etal-1993-building} and Chinese Treebank (CTB) 5.1 \cite{ctb} according to standard data splitting (39832/1700/2416 for PTB and 17544/352/348 for CTB). 
Besides, we test the generalization ability of our model on Multi-domain Constituency Treebank (MCTB) \cite{yang-etal-2022-challenges}.
We calculate F1 score for constituency parsing by using the standard EVALB tool \cite{roark-etal-2006-sparseval}.

\subsection{Setup}
We employ the pre-trained models BERT-large-uncased for English and BERT-base-chinese for Chinese \cite{devlin-etal-2019-bert,wolf-etal-2020-transformers}.  
For the span scorer, we utilize a one-layer feed-forward neural network with a 250-dimensional hidden layer. 
The encoder is set to the same as \citet{mrini-etal-2020-rethinking} with 12 self-attention layers. 
The dimension of query, key, value and output vectors all have 128 dimensions for both languages. The random seed is set to 777.

\begin{table}[t]
\begin{center}
\resizebox{\linewidth}{!} {
\begin{tabular}{cccc}
\hline \bf $\lambda$ & \bf LR & \bf LP & \bf F1 \\\hline
0.0 & 94.86 $\pm$ 0.07 & 95.07 $\pm$ 0.05 & 94.96 $\pm$ 0.06\\
0.1 & 94.89 $\pm$ 0.05 & 95.09 $\pm$ 0.06 & 94.99 $\pm$ 0.05\\
0.2 & 95.05 $\pm$ 0.05 & 95.14 $\pm$ 0.04 & 95.09 $\pm$ 0.04\\
0.3 & 95.01 $\pm$ 0.04 & 95.13 $\pm$ 0.04 & 95.07 $\pm$ 0.04\\
0.4 & 95.15 $\pm$ 0.04 & \textbf{95.23} $\pm$ 0.02 & \textbf{95.19} $\pm$ 0.03\\
0.5 & 95.12 $\pm$ 0.05 & 95.06 $\pm$ 0.04 & 95.09 $\pm$ 0.04\\
0.6 & 94.93 $\pm$ 0.05 & 95.18 $\pm$ 0.05 & 95.06 $\pm$ 0.05\\
0.7 & \textbf{95.17} $\pm$ 0.03 & 95.05 $\pm$ 0.04 & 95.11 $\pm$ 0.04\\
0.8 & 95.05 $\pm$ 0.04 & 95.03 $\pm$ 0.03 & 95.04 $\pm$ 0.03\\
0.9 & 95.06 $\pm$ 0.05 & 94.96 $\pm$ 0.04 & 95.01 $\pm$ 0.05\\
1.0 & 95.01 $\pm$ 0.03 & 94.81 $\pm$ 0.03 & 94.91 $\pm$ 0.03\\\hline
\end{tabular}
}
\end{center}
\caption{Ranging $\lambda$ from 0 to 1 on the PTB dev set.}
\label{tab:combine} 
\end{table}

\begin{table}[t]
\begin{center}
\resizebox{\linewidth}{!}{
\begin{tabular}{lllll}
\hline Node & Model & LR & LP & F1 \\\hline
\multirow{3}*{\$+label} & Baseline & 95.34 & 94.35 & 94.84\\
& Ours ($\lambda=0.0$) & \bf 95.48 & 94.72 & 95.10\\
& Ours ($\lambda=0.4$) & 95.42 & 94.62 & 95.02\\\hline
\multirow{3}*{\$} & Baseline & 94.62 & 95.01 & 94.81\\
& Ours ($\lambda=0.0$) & 94.86 & 95.07 & 94.96\\
& Ours ($\lambda=0.4$) & 95.15 & \bf 95.23 & \bf 95.19\\\hline
\end{tabular}
}
\end{center}
\caption{Results of two labeling schemes (Section \ref{BBB}) on the PTB dev set.}
\label{tab:morerules} 
\end{table}

\begin{table}[t]
\begin{center}
\resizebox{\linewidth}{!}{
\begin{tabular}{lccc}
\hline
\bf Model & \bf LR & \bf LP & \bf F1 \\\hline
\citet{nguyen-etal-2020-efficient} & - & - & 95.48 \\
\citet{jiang-etal-2020-generalizing} & - & - & 95.5 \\
\citet{kitaev-etal-2019-multilingual} & 95.46 & 95.73 & 95.59 \\
\citet{baevski-etal-2019-cloze} & - & - & 95.6 \\
\citet{YuZhang2020FastAA} & 95.55 & 95.85 & 95.69\\
\citet{chen-etal-2021-neural} & - & - & 95.72\\
\citet{zhou-zhao-2019-head} & 95.51 & 95.93 & 95.72 \\
\citet{zhou-zhao-2019-head}* & 95.70 & 95.93 & 95.84 \\
\citet{wei-etal-2020-span} & 95.5 & 96.1 & 95.8\\
\citet{tian-etal-2020-improving} & 95.58 & 96.11 & 95.85 \\
\citet{xin-etal-2021-n} & 95.55 & \bf 96.29 & \bf 95.92 \\\hline
Baseline & 95.38 & 95.84 & 95.61\\
Ours (w/o rule inference) & 95.48 & 95.89 & 95.68\\
Ours & \bf 95.71 & 96.08 & 95.89 \\\hline
\end{tabular}
}
\end{center}
\caption{Performance on the PTB test set. * indicates training with extra dependency parsing data.}
\label{tab:pp1} 
\end{table}

\begin{table}[t]
\begin{center}
\resizebox{\linewidth}{!}{
\begin{tabular}{lccc}
\hline
\bf Model & \bf LR & \bf LP & \bf F1 \\\hline
\citet{kitaev-etal-2019-multilingual} & 91.55 & 91.96 & 91.75\\
\citet{zhou-zhao-2019-head} & 91.14 & \bf 93.09 & 92.10\\
\citet{tian-etal-2020-improving} & 92.14 & 92.25 & 92.20 \\
\citet{YuZhang2020FastAA} & 92.04 & 92.51 & 92.27\\
\citet{wei-etal-2020-span} & 92.2 & 92.7 & 92.4\\
\citet{xin-etal-2021-n} & 92.06 & 92.94 & 92.50 \\\hline
Baseline & 91.82 & 92.28 & 92.05\\
Ours (w/o rule inference) & 91.97 & 92.33 & 92.15\\
Ours & \bf 92.22 & 92.82 & \bf 92.52\\\hline
\multicolumn{4}{c}{w/ External Dependency Parsing Data} \\\hline
\citet{zhou-zhao-2019-head}* & 92.03 & 92.33 & 92.18\\
\citet{mrini-etal-2020-rethinking}* & 91.85 & 93.45 & 92.64\\\hline
\end{tabular}
}
\end{center}
\caption{Performance on the CTB test set. * indicates training with extra dependency parsing data.}
\label{tab:pp2} 
\end{table}

\begin{table*}[t]
\begin{center}
\resizebox{\linewidth}{!}{
\begin{tabular}{c|c|ccccc|c}
\hline
\multirow{2}*{\bf Model} & \bf In-domain & \multicolumn{6}{c}{\bf Cross-domain} \\
\cline{2-8} 
& \bf PTB & \bf Dialogue & \bf Forum & \bf Law & \bf Literature & \bf Review & \bf Avg \\\hline
\citet{liu-zhang-2017-shift} & 95.65 & 85.56 & 86.33 & 91.50 & 84.96 & 83.89 & 86.45\\
\citet{kitaev-klein-2018-constituency} & 95.73 & 86.30 & 87.04 & 92.06 & 86.26 & 84.34 & 87.20\\\hline
Ours (w/o rule inference) & 95.68 & 85.74 & 86.28 & 91.68 & 85.20 & 84.56 & 86.68\\
Ours & \bf 95.89 & \bf 87.16 & \bf 87.56 & \bf 92.83 & \bf 86.95 & \bf 85.96 & \bf 88.09\\\hline
\end{tabular}
}
\end{center}
\caption{Cross-domain performance on MCTB (1000 sentences for each domain). \citet{liu-zhang-2017-shift} and \citet{kitaev-klein-2018-constituency} are reported by \citet{yang-etal-2022-challenges}.}
\label{tab:cross-domain} 
\end{table*}

\subsection{Joint Training}
In our joint training, the weight $\lambda$ is a hyperparameter that is crucial in balancing the role of two different parsers.
We conduct each experiment 5 times with different $\lambda$ and report the mean in Table \ref{tab:combine}.
It shows that our joint training outperforms either rule-less training or rule-based training.
A single conventional CKY parser predicts the label of each span precisely but ignores potential syntactic relations between spans, which fails to handle complex semantics. 
On the other hand, a single rule-based CKY parser is hard to train, and we hypothesize that it is because the rule-based CKY decoder prevents the model from learning from errors made by the conventional CKY parser. Since neural networks are trained from the back-propagation error, excluding errors during training might harm the model from understanding the overall grammar.
As shown in Table \ref{tab:combine}, our model achieves the best result when $\lambda$ is set to 0.4.

\subsection{Different Labeling Schemes}
\label{sec:com}
In Section \ref{BBB} we mention that we try to denote the labels of generated nodes using two labeling schemes.
The comparison results are shown in Table \ref{tab:morerules}, and we obtain the better result using the first one, where the same label \$ is used for all generated nodes.
Although the baseline (without rules) of the two approaches are very close, using \$+label for denotation obtains the best result only when the rule-based CKY decoder is used for inference, and with a mixture of rule-based training and rule-less training, it is even less effective than rule-less training.
These results suggest that allowing the model to learn excessive rules with more labels can be less effective.
Using \$ to denote is more straightforward and effective.

\subsection{Main Results}
In Table \ref{tab:pp1} and Table \ref{tab:pp2}, we compare the performance of our rule-based parser with previous BERT-based parsers on PTB and CTB, respectively. 
\citet{zhou-zhao-2019-head} use head-driven phrase structure grammar to employ multi-task training of constituency parsing and dependency parsing, and \citet{mrini-etal-2020-rethinking} further extend it with the label attention layer.
In our baseline, we reproduce the prior state-of-the-art parser \cite{mrini-etal-2020-rethinking} without using external dependency parsing data. 
On the PTB and CTB test sets, our model achieves 95.89 and 92.52 F1 respectively, which shows an absolute improvement of 0.28 and 0.47 points compared with our baselines.

Compared with other single-task models, our result on PTB outperforms most of them. The best result \cite{xin-etal-2021-n} utilizes the sibling dependencies of $n$-ary nodes' children, rather than applying the CKY algorithm into binary trees.
Our result on CTB outperforms all previous state-of-the-art parsers, and is even competitive with the result of \citet{mrini-etal-2020-rethinking} that uses multi-task training with extra dependency parsing data.
In summary, our results demonstrate the advantage of incorporating syntactic rules into decoding and the joint training of a rule-based parser and a rule-less parser.

\begin{figure}[t]
\centering
\includegraphics[width=\linewidth]{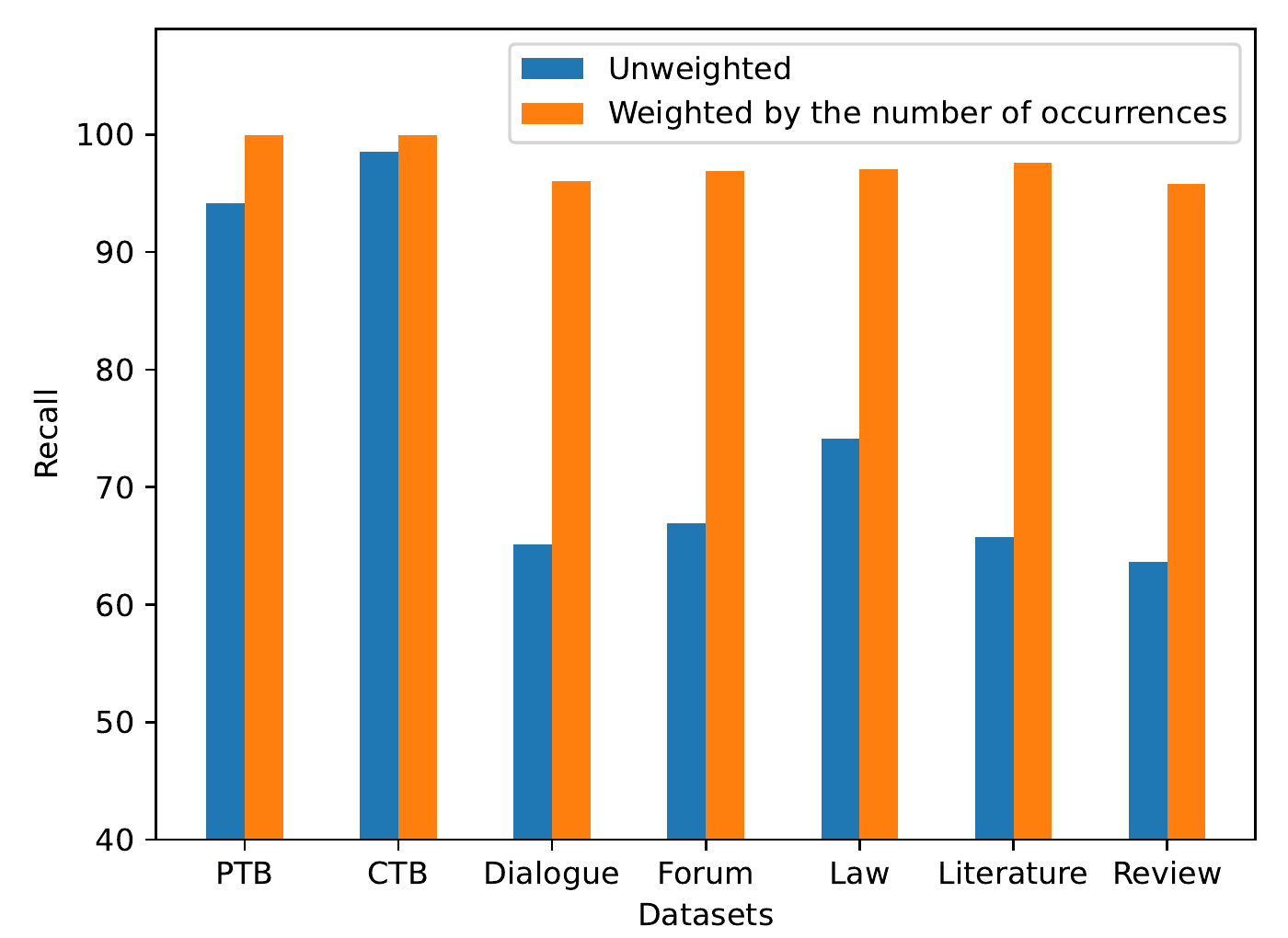}
\caption{Statistics of the number of rules in each test set compared to the training set.}
\label{fig:recall}
\end{figure}

\begin{table}[t]
\begin{center}
\resizebox{\linewidth}{!}{
\begin{tabular}{lr}
\hline
\bf Model & \bf sents/sec\\\hline
\citet{shen-etal-2018-straight} & 111\\
\citet{shen-etal-2018-straight} (w/o tree inference) & 351\\
\citet{zhou-zhao-2019-head} & 65\\
\citet{gnn} & 71\\
\citet{mrini-etal-2020-rethinking} & 60\\
\citet{xin-etal-2021-n} & 26\\\hline
Ours (w/o rule inference) & 121\\
Ours & 54\\\hline
\end{tabular}
}
\end{center}
\caption{Speed comparison of different parsers using different hardware on the PTB test set. Rows 4-6 are reported by \citet{gnn} using a GeForce GTX 2080 Ti. \citet{xin-etal-2021-n} use a GeForce RTX 3090, and we use a GeForce Titan RTX.}
\label{tab:speed2} 
\end{table}

\subsection{Cross-domain Results}
In Table \ref{tab:cross-domain}, we compare the generalization ability of our method with previous approaches.
All the parsers are trained on the PTB training set and evaluated on the PTB dev set, and tested their cross-domain performance on MCTB \cite{yang-etal-2022-challenges} in zero-shot settings.
Our rule-based parser achieves 88.09 F1 on average in cross-domain datasets with an absolute improvement of 1.41 points compared with our rule-less parser, which also outperforms other previous approaches.

To further explore the role of syntactic rules in our parser, we count the number of rules in each test set compared to the rules in the training set (the PTB training set for English and the CTB training set for Chinese).
As shown in Figure \ref{fig:recall}, our syntactic rules in both PTB and CTB test sets have over 99.9\% recalls (weighted by the number of occurrences) over their training sets. 
In cross-domain datasets, although about 25\% to 35\% of the rules do not appear in the PTB training set, the weighted statistics show that the recalls of the rules in cross-domain datasets are all over 95.8\%.
This demonstrates that our simple rule labeling scheme extends effectively to other domain datasets, which also improves the generalization ability of our model.
Although our data-driven algorithm cannot generalize to a small proportion of unseen rules without a specific generalization objective, our method can utilize known syntactic rules to correct errors, resulting in better parsing performance.

\subsection{Speed Analysis}
Our method can be completely parallelized using GPU accelerations. 
For each epoch, it takes 15.2 minutes for rule-less training,  
and 17.5 minutes under a mixture of rule-based training and rule-less training,
which is merely increased by 15\%. It shows the efficiency of our GPU accelerated implementation.

We compare our parsing speed with other parsers in Table \ref{tab:speed2}, although this comparison might be unfair since they are run on different hardware.
Our rule-less parser processes 121 sentences per second, and our rule-based parser processes 54 sentences per second.
Compared with \citet{xin-etal-2021-n} which use a GeForce RTX 3090, our parser on a GeForce Titan RTX is much faster, while F1 scores of the two parsers are comparable.

\section{Conclusion}
\label{sec:conclusion}
In this paper, we first implement a fast neural CKY decoding, based on which we further propose a rule-based CKY decoding, whose time complexity is $O(n)$ in terms of GPU operations.
We also investigate the joint training of a rule-based parser and a conventional rule-less parser.
Our experiments show that incorporating syntactic rules into decoding and training can rectify errors in the parse tree, which achieves significant improvements on both English and Chinese benchmark treebanks. 
Furthermore, we demonstrate that our method is robust in zero-shot cross-domain settings.

\bibliography{anthology,custom}
\bibliographystyle{acl_natbib}

\appendix

\end{document}